\documentclass[letterpaper, 10 pt, conference]{ieeeconf}  

\IEEEoverridecommandlockouts                              

\usepackage{amsmath} 
\usepackage{amssymb}  
\usepackage{graphicx}

\usepackage{stfloats}    

\usepackage{placeins} 

\usepackage[caption=false,font=footnotesize]{subfig} 

\title{\LARGE \bf
Sensorless Four-Channel Control Architecture Using Inverse Dynamics Modeling for Human-Scale Bilateral Teleoperation
}

\author{Amir Noohian$^{1*}$, Dylan Miller$^{2}$, Justin Valentine$^{2}$, Alan Lynch$^{1}$ and Martin Jagersand$^{2}$
\thanks{*Corresponding author}
\thanks{$^{1}$Amir Noohian and Alan Lynch are with the Department of Electrical and Computer Engineering, University of Alberta, Canada.
        {\tt\small \{noohian; alan.lynch\}@ualberta.ca}}%
\thanks{$^{2}$Dylan Miller, Justin Valentine and Martin Jagersand are with the Department of Computing Science, University of Alberta, Canada.
        {\tt\small \{djm2;jvalenti;mj7\}@ualberta.ca}}%
}

\begin{document}

\maketitle
\thispagestyle{empty}
\pagestyle{empty}

\begin{abstract}

The four-channel teleoperation architecture is a well-established framework for achieving transparency in bilateral systems. However, its performance in human-scale teleoperation is limited by high inertia, modeling challenges, and reliance on noisy and costly force/torque sensors. This paper introduces a sensorless four-channel architecture based on inverse dynamics modeling. The controller is implemented and validated on a customized WAM bilateral teleoperation setup. Experiments demonstrate that the proposed approach outperforms conventional two- and four-channel schemes as well as transparency-enhancement methods, improving position and force tracking, reducing operator effort, and increasing maximum transmittable impedance without external sensors. A door-opening case study involving sustained whole-body contact along the manipulator further demonstrates the effectiveness of the method in realistic human-scale manipulation tasks.

\end{abstract}

\section{INTRODUCTION}

Bilateral teleoperation systems enable operators to perform complex tasks in remote or hazardous environments with greater precision and safety \cite{maeso2010efficacy, fong2013space, khasawneh2019human}. An ideal system allows the operator, through a haptic interface (leader), to feel as if directly interacting with the remote environment. This is achieved by accurately reproducing the follower–environment interaction forces, allowing clear perception of object locations, surface properties, and contact forces for precise and reliable control \cite{laghi2020unifying}. Haptic interfaces range from lightweight devices such as the Phantom Omni to human-scale systems like the DLR teleoperation facility HUG \cite{hulin2011dlr}. While smaller devices offer portability, human-scale systems provide larger workspaces and higher force capabilities, improving dexterity and interaction realism \cite{glover2009effective}.

The key performance objective in human-scale teleoperation is transparency, meaning the operator perceives the remote environment without distortion, as if connected through a massless and infinitely stiff link. Achieving transparency requires continuous exchange of position and force information between the leader and follower, along with accurate compensation of robot dynamics \cite{lawrence1993stability}. In practice, however, this is difficult to realize. Human-scale manipulators are massive and introduce significant damping, increasing operator effort. Moreover, many transparent control schemes rely on force/torque sensors that are costly, noise-sensitive, difficult to integrate \cite{su2020deep}, and typically limited to end-effector measurements, preventing detection of interactions along the manipulator body. These limitations restrict high-fidelity force rendering and make stable, transparent bilateral control particularly challenging.

\begin{figure}[t]
  \centering
  \subfloat[]{\includegraphics[width=0.45\columnwidth]{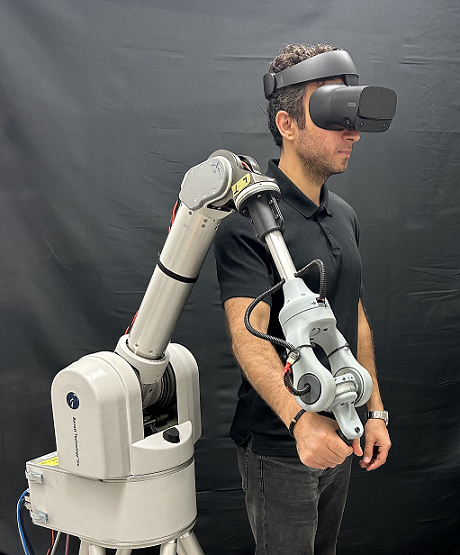}\label{fig:wam_leader}}
  \hspace{0.02\columnwidth}
  \subfloat[]{\includegraphics[width=0.45\columnwidth]{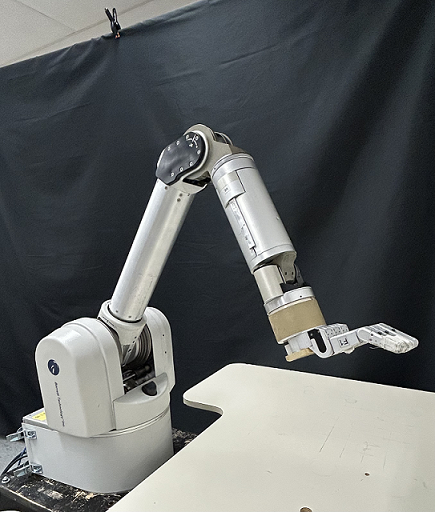}\label{fig:wam_follower}}
  \caption{WAM bilateral teleoperation system setup: (a) 4-DOF leader arm with custom haptic wrist, (b) 7-DOF follower arm.}
  \label{fig:wam teleop setup}
  \vspace{-8pt}
\end{figure}

To address these challenges, this paper proposes a four-channel control architecture implemented on a human-scale bilateral WAM teleoperation system (Fig.~\ref{fig:wam teleop setup}). The approach leverages inverse dynamics modeling and eliminates the need for force/torque sensors, enabling transparent teleoperation while reducing system complexity and cost.

\section{RELATED WORK}

\subsection{Inverse Dynamics Estimation}

Accurate dynamic models are essential for implementing model-based controllers. A widely used approach to obtain inverse dynamics is robot parameter estimation, where dynamic parameters are identified using analytical \cite{gautier1990direct} or numerical \cite{gautier1991numerical} methods. Several studies have improved this approach by enforcing physical feasibility of the estimated parameters \cite{sousa2013physically} or considering the manipulator mounting configuration \cite{trobinger2023identification}.

Inverse dynamics can also be learned using data-driven methods that require little or no prior knowledge of the robot model. These include non-parametric approaches such as locally weighted projection regression (LWPR) and Gaussian process regression (GPR) \cite{nguyen2008computed, nguyen2009model}, as well as parametric methods like artificial neural networks (ANN) \cite{morse2020learning, hitzler2019learning}. More recent works incorporate physical priors into learning frameworks, such as rigid-body kernels in GPR \cite{nguyen2010using}, deep Lagrangian networks (DeLaN) \cite{lutter2019deep}, and physics-informed neural networks (PINN) \cite{yang2023physics}.

Compared to learning-based approaches, model-based parameter estimation requires fewer parameters and directly exploits the robot’s physical structure, leading to improved generalizability and robustness. Moreover, the resulting dynamics computation incurs negligible latency, which is critical for real-time bilateral teleoperation. Therefore, we adopt a parameter estimation approach to obtain the inverse dynamics of the leader and follower robots in our WAM teleoperation system.

\subsection{Transparency in Teleoperation Systems}

Improving transparency in bilateral teleoperation has been a central research focus for decades. Early studies such as \cite{lawrence1993stability} analyzed teleoperation architectures and emphasized the importance of dynamic compensation for achieving high transparency. Since then, several approaches have been proposed to enhance performance, including inverse dynamics for impedance control \cite{tufail2015haptic}, adaptive control frameworks \cite{5979591}, and nonlinear disturbance observers \cite{7982884}. While these methods improved stability and transparency, most were validated only in simulations or simple one-degree-of-freedom (1-DOF) setups, limiting their applicability to real-world, human-scale systems.

Recent works in telesurgery have applied learning-based techniques for inverse dynamics and force estimation. Deep neural networks were used on the da Vinci Research Kit to identify inverse dynamics and external forces \cite{yilmaz2020neural}, later extended to a sensorless four-channel teleoperation scheme combining dynamics compensation and disturbance observers \cite{yilmaz2023sensorless}. Although superior tracking was reported, leader-side impedance was not evaluated, which is a key indicator of how naturally the operator perceives interaction.

In human-scale teleoperation, transparency is further challenged by the high inertia and damping of longer manipulators. Force feedforward improved free-motion transparency \cite{fahmi2018inertial}, and closed-loop force control enhanced force tracking on the DLR-HUG platform \cite{balachandran2020closing}, but both remain limited in hard contact, particularly in achieving high maximum transmittable impedance. Moreover, these approaches rely on costly end-effector force/torque sensors, which cannot detect contacts along the robot body. This is a significant drawback for large systems, where tasks often involve body contact and collisions, leaving such interactions invisible to end-tip sensing and limiting both safety and transparency.

\subsection{Contributions}

We propose a sensorless four-channel teleoperation architecture that employs inverse dynamics modeling to estimate external joint torques without force/torque sensors. The architecture integrates feedforward dynamic compensation on both the leader and follower and uses the estimated interaction torques to provide haptic feedback to the operator. The main contributions of this work are:

\begin{itemize}
\item Development and implementation of a sensorless four-channel teleoperation architecture for human-scale manipulators.
\item Real-time estimation of external joint torques via inverse dynamics, enabling detection of contact along the manipulator body rather than only at the end-effector.
\item Experimental validation and objective transparency evaluation on a customized WAM bilateral teleoperation system (Fig.~\ref{fig:wam teleop setup}).
\item Comparative analysis against conventional two-channel, four-channel, and transparency-enhancement teleoperation approaches.
\end{itemize}

\section{METHODOLOGY}

\subsection{General Transparency Analysis Framework}\label{transparencyanalysis}

A teleoperation system can be represented using a two-port model. For a linearized 1-DOF system in the frequency domain,

\begin{equation}\label{eq:1}
\begin{aligned}
    Z_lX_l &= F_{h} + F_{l},\\
    Z_fX_f &= -F_{e} + F_{f},
\end{aligned}
\end{equation}
where $Z_l$ and $Z_f$ denote the leader and follower impedances, $X_l$ and $X_f$ their positions, $F_h$ and $F_e$ the human and environment forces, and $F_l$, $F_f$ the control inputs. Transparency can be quantified through the hybrid matrix $H$,

\begin{equation}\label{eq:2}
\begin{bmatrix}
F_h \\
X_f
\end{bmatrix}
=
\begin{bmatrix}
h_{11} & h_{12} \\
h_{21} & h_{22}
\end{bmatrix}
\begin{bmatrix}
X_l \\
F_e
\end{bmatrix}.
\end{equation}

Perfect transparency is achieved when the transmitted impedance $Z_{to}=F_h/X_l$ equals the environment impedance $Z_e=F_e/X_f$. Expressing $Z_{to}$ in terms of $Z_e$ yields

\begin{equation}\label{eq:4}
    Z_{to} = \frac{h_{11} + Z_e \left(h_{12}h_{21} - h_{11}h_{22}\right)}{1 - h_{22} Z_e}.
\end{equation}
According to \cite{hannaford1989design}, full transparency requires

\begin{equation}\label{eq:5}
    H =
    \begin{bmatrix}
        0 & 1\\
        1 & 0
    \end{bmatrix}.
\end{equation}

Here, $h_{11}$ represents leader impedance and $h_{21}$ free-motion position tracking. Since $h_{12}$ and $h_{22}$ correspond to leader hard contact, which is not practical, we instead use the experimentally measurable parameters \cite{aliaga2004experimental}

\begin{equation} \label{eq:6}
    F_{12} = \frac{F_h}{F_e}\big|_{X_f = 0}, \quad
    Z_{11} = \frac{F_h}{X_l}\big|_{X_f = 0},
\end{equation}
where $F_{12}$ denotes hard-contact force tracking and $Z_{11}$ the maximum transmittable impedance. Together with $h_{11}$ and $h_{21}$, these metrics characterize transparency and can be evaluated experimentally.

\subsection{Four-Channel Teleoperation System Architecture}\label{fourchannel}

\begin{figure}[t]
    \centering  \includegraphics[width=\columnwidth]{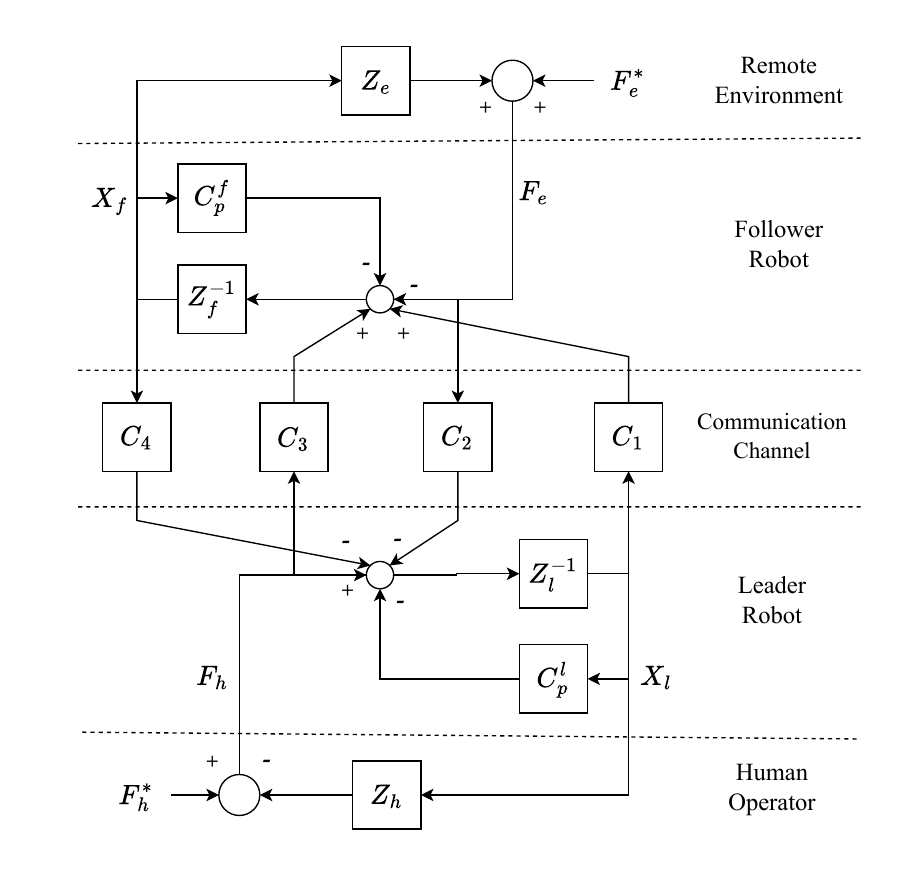}
    \caption{Four-channel teleoperation system block diagram.}
    \label{fig:fourchannel block diagram}
    \vspace{-15pt}
\end{figure}

According to \cite{lawrence1993stability}, the transparency in a teleoperation system can be optimized when both force and position measurements are exchanged between the robots and the robot dynamics are perfectly compensated. Fig. \ref{fig:fourchannel block diagram} shows the block diagram of the transparency-optimized teleoperation system, which is composed of four channels of communication: \(C_1\) and \(C_4\) for position and \(C_2\) and \(C_3\) for force.

The control efforts for leader and follower robots can be written as

\begin{equation}\label{eq:7}
\begin{aligned}
F_{l} &= -C_4 X_f - C_p^l X_l - C_2 F_e,\\
F_{f} &= C_1 X_l - C_p^f X_f + C_3 F_h,
\end{aligned}
\end{equation}
where the position and force channels can be designed as

\begin{equation}\label{eq:8}
\begin{aligned}
    C_1 &= Z_f + C_p^f,\\
    C_2 &= C_f^l,\\
    C_3 &= C_f^f,\\
    C_4 &= -(Z_l + C_p^l),
\end{aligned}
\end{equation}
where \({C_p^f} = k_p^f + \frac{k_i^f}{s} + k_d^fs\) and \({C_p^l} = k_p^l + \frac{k_i^l}{s} + k_d^ls\) are PID position controllers and \(C_f^f = k_f^f\) and \(C_f^l = k_f^l\) are force feedback gains for the follower and leader robots, respectively. Substituting (\ref{eq:8}) in (\ref{eq:7}), we can write the controllers as

\begin{equation}
\begin{aligned}
F_{l} &= Z_l X_f + C_p^l (X_f - X_l) - C_f^l F_e,\\
F_{f} &= Z_f X_l + C_p^f (X_l - X_f) + C_f^f F_h,
\end{aligned}
\label{eq:9}
\end{equation}

The controllers are composed of three parts: the first term compensates for the robot impedance/dynamics, the second term provides the PID feedback of position error, and the third term is the force feedback. We select \(C_p^l = C_p^f = C_p\) and \(C_f^l = C_f^f = C_f\) for identical follower and leader robots.

Assuming perfect compensation of the robot dynamics and no force feedback ($C_f=0$) yields a two-channel position--position (P--P) controller with dynamics compensation. The transparency parameters can be obtained as
\begin{equation}\label{eq:10}
    \begin{aligned}
        h_{11} = 0,
        h_{21} = 1,
        F_{12} = 1,
        Z_{11} = Z_l+C_p,
    \end{aligned}
\end{equation}
which gives the optimal transparency for free motion position tracking and leader impedance, as well as hard contact force tracking. However, it does not provide the ideal maximum transmittable impedance. 

Using force feedback to have a four-channel teleoperation system with dynamic compensation, the maximum transmittable impedance is obtained as follows

\begin{equation}
    \begin{aligned}
    \label{eq:11}
        Z_{11} &= \frac{(Z_l+C_p) + (Z_f + C_p) C_f}{1-C_f^2},
    \end{aligned}
\end{equation}
which shows that the ideal maximum transmittable impedance can be achieved by setting \(C_f=1\). In fact, as shown in (\ref{eq:10}) and (\ref{eq:11}), although dynamics compensation can help with free motion transparency, the use of force feedback is necessary to achieve the optimal transparency in hard contact.

\subsection{Inverse Dynamics Modeling for Manipulators}\label{inversedynamics}

A robot manipulator is modeled as an open kinematic chain of \(n+1\) rigid bodies and \(n\) joints. The generalized coordinates \(q \in \mathbb{R}^n\) represent the joint angles. The robot dynamics are given by
\begin{equation}
\label{eq:12}
M(q)\ddot{q} + b(q, \dot{q})\dot{q} + g(q) + \tau_d = \tau,
\end{equation}
where \(M(q) \in \mathbb{R}^{n \times n}\) is the inertia matrix, \(b(q, \dot{q})\dot{q} \in \mathbb{R}^n\) the Coriolis and centrifugal torque, \(g(q) \in \mathbb{R}^n\) the gravity torque, \(\tau \in \mathbb{R}^n\) the actuation torque, and \(\tau_d \in \mathbb{R}^n\) the dissipative torque, mainly due to joint friction.

Several friction models were evaluated 
\cite{desai2001towards, colome2015friction, huang2025adaptive}, 
but none provided satisfactory compensation. Accurately capturing static and Coulomb friction proved difficult, often leading to instability and loss of passivity. Therefore, only viscous friction was retained, yielding
\(\tau_d = F_v \dot{q}\), where \(F_v \in \mathbb{R}^{n \times n}\) is the diagonal matrix of viscous friction coefficients. The dynamic model in \eqref{eq:12} can be linearized with respect to a set of inertial parameters as
\begin{equation} \label{eq:14}
\underbrace{
\begin{bmatrix}
\tau_1 \\
\tau_2 \\
\vdots \\
\tau_n
\end{bmatrix}
}_{\tau}
=
\underbrace{
\begin{bmatrix}
y_{11}^\top & y_{12}^\top & \cdots & y_{1n}^\top \\
0 & y_{22}^\top & \cdots & y_{2n}^\top \\
\vdots & \vdots & \ddots & \vdots \\
0 & 0 & \cdots & y_{nn}^\top
\end{bmatrix}
}_{Y(q, \dot{q}, \ddot{q})}
\underbrace{
\begin{bmatrix}
\pi_1 \\
\pi_2 \\
\vdots \\
\pi_n
\end{bmatrix}
}_{\pi},
\end{equation}
where \( Y(q, \dot{q}, \ddot{q}) \in \mathbb{R}^{n \times L} = \frac{\partial \tau}{\partial \pi} \) is the regressor matrix, computed from joint position \( q \), velocity \( \dot{q} \), and acceleration \( \ddot{q} \). The inertial parameters of link \( i \) are
\begin{equation} \label{eq:15}
\begin{aligned}
\pi_i = [L_{xx,i}, L_{xy,i}, L_{xz,i}, L_{yy,i}, L_{yz,i}, L_{zz,i}, \\
l_{x,i}, l_{y,i}, l_{z,i}, m_i, F_{vi}]^\top,
\end{aligned}
\end{equation}
where \( L_{xx,i}, \ldots, L_{zz,i} \) are the inertia tensor components in frame \( i \), \( m_i \) is the mass, \( F_{vi} \) the viscous friction coefficient, and \( l_{x,i}, l_{y,i}, l_{z,i} \) are the first moments of mass,
\begin{equation}\label{eq:16}
[l_{x,i}, 
l_{y,i},
l_{z,i}]^\top
= m_i r_{c,i},
\end{equation}
with \( r_{c,i} \in \mathbb{R}^3 \) the center of mass in frame \( i \). Only a subset of these parameters is identifiable. Analytical \cite{gautier1990direct} or numerical \cite{gautier1991numerical} methods are used to obtain the base inertial parameters \cite{gautier1990direct}. The dynamics can then be written as
\begin{equation} \label{eq:17}
\tau = Y_b(q, \dot{q}, \ddot{q}) \pi_b,
\end{equation}
where \(Y_b\) is the regressor for the base parameters and  \(\pi_b \in \mathbb{R}^b\) the base parameter vector. Given an excitation trajectory with \(N>b\) samples, the overdetermined system
\begin{equation} \label{eq:18}
\bar{\tau} = \bar{Y}_b \pi_b
\end{equation}
is obtained, and the least-squares estimate is
\begin{equation} \label{eq:19}
\hat{\pi}_b = (\bar{Y}_b^\top \bar{Y}_b)^{-1} \bar{Y}_b^\top \bar{\tau}.
\end{equation}

To ensure parameter excitation and avoid rank deficiency of \(\bar{Y}_b\), joint trajectories are modeled as finite Fourier series,
\begin{equation} \label{eq:20}
q_i(t) = q_{i0} + \sum_{k=1}^{M} \left( a_{i,k} \sin(k \omega_f t) + b_{i,k} \cos(k \omega_f t) \right),
\end{equation}
where \(q_{i0}\) is the offset and \(\omega_f\) the fundamental frequency with period \(T_f = 2\pi/\omega_f\). The coefficients \(a_{i,k}\) and \(b_{i,k}\) are chosen to minimize the condition number of the regressor matrix.

\subsection{External Torque Estimation}\label{externaltorque}

When the manipulator is in contact with the environment, the dynamic model in (\ref{eq:12}) becomes

\begin{equation}
\label{eq:21}
M(q)\ddot{q} + b(q, \dot{q})\dot{q} + g(q) + \tau_d + \tau_{ext}= \tau,
\end{equation}
where \(\tau_{ext}\) denotes the external torque applied at the joints. Replacing the true dynamics with the estimated dynamics from (\ref{eq:19}) yields

\begin{equation}
\label{eq:22}
Y_b(q, \dot{q}, \ddot{q}) \hat{\pi}_b + \tau_{ext}= \tau.
\end{equation}
Thus, the external torque is estimated as

\begin{equation}
\label{eq:23}
\hat{\tau}_{ext} = \tau - Y_b(q, \dot{q}, \ddot{q}) \hat{\pi}_b .
\end{equation}

The external Cartesian wrench is related to the external joint torques through the manipulator Jacobian as

\begin{equation}
\label{eq:222}
\tau_{ext} = J^T F_{ext},
\end{equation}
where $J \in \mathbb{R}^{6 \times n}$ is the Jacobian and 
$F_{ext} \in \mathbb{R}^{6}$ is the external Cartesian wrench.

\subsection{Sensorless Four-Channel Control on the WAM Teleoperation System}\label{transparencyoptimizedwam}

In a teleoperation system, the dynamic model of the leader and follower robots in the joint space can be written as
\begin{equation}\label{eq:24}
\begin{aligned}
M_l(q_l)\ddot{q}_l + b_l(q_l, \dot{q}_l)\dot{q}_l + g_l(q_l) + {\tau_d}_l &= {{\tau}}_l + \tau_h,\\
M_f(q_f)\ddot{q}_f + b_f(q_f, \dot{q}_f)\dot{q}_f + g_f(q_f) + {\tau_d}_f &= {{\tau}}_f - \tau_e,
\end{aligned}
\end{equation}
where ${{\tau}}_l, {{\tau}}_f \in \mathbb{R}^n$ are leader and follower control efforts, and $\tau_h, \tau_e \in \mathbb{R}^n$ are the human-applied torque and the external environmental torque, respectively.

To implement the four-channel controller described in Section \ref{fourchannel} in joint space, we can write the control efforts as

\begin{equation}\label{eq:25}
\begin{aligned}
{{\tau}}_l &= {\tau_\textit{ff}}_l + {C}_p^l(q_f - q_l) - C_f^f \hat{\tau}_{e} ,\\
{{\tau}}_f &= {\tau_{\textit{ff}}}_f + {C}_p^f(q_l - q_f) + C_f^l \hat{\tau}_h,
\end{aligned}
\end{equation}
where \({\tau_\textit{ff}}_l, {\tau_\textit{ff}}_f \in \mathbb{R}^n\) are the leader and follower feedforward terms, respectively. 
Using the estimated inverse dynamics described in Section \ref{inversedynamics} as the feedforward term, we have

\begin{equation}\label{eq:26}
\begin{aligned}
{{\tau}}_l &= Y_{b_l}(q_f, \dot{q}_f, \ddot{q}_f)\hat{\pi}_{b_l} + {C}_p^l(q_f - q_l) - C_f^f \hat{\tau}_{e} ,\\
{{\tau}}_f &= Y_{b_f}(q_l, \dot{q}_l, \ddot{q_l})\hat{\pi}_{b_f} + {C}_p^f(q_l - q_f) + C_f^l \hat{\tau}_h,
\end{aligned}
\end{equation}
where \(\hat{\pi}_{b_l}, \hat{\pi}_{b_f} \in \mathbb{R}^b\) are the estimated dynamic parameter vectors, \(Y_{b_l}, Y_{b_f}\) are the regressor matrices. $C_p^l, C_p^f \in \mathbb{R}^{n \times n}$ are diagonal matrices with $C_{p,i}^l, C_{p,i}^f$ as PID joint position controllers on the diagonals, and $C_{f}^l,C_{f}^f \in \mathbb{R}^{n \times n}$ are diagonal matrices with $C_{f,i}^l,C_{f,i}^f$ as joint torque feedback gains on the diagonals, for the leader and follower robots, respectively. Also, \(\hat{\tau}_{e}, \hat{\tau}_h \in \mathbb{R}^n\) are the estimated external environmental and human-applied torques, which can be estimated by applying the external torque estimation method described in Section \ref{externaltorque} to (\ref{eq:24}) as follows

\begin{equation} \label{eq:27}
\begin{aligned}
\hat{\tau}_{e} &= \tau_f - Y_{b_f}(q_f, \dot{q}_f, \ddot{q}_f)\hat{\pi}_{b_f},\\
\hat{\tau}_{h} &= Y_{b_l}(q_l, \dot{q}_l, \ddot{q}_l)\hat{\pi}_{b_l} - \tau_l.
\end{aligned}
\end{equation}

\begin{figure}[t]
        \centering
        \includegraphics[width=0.5\textwidth]{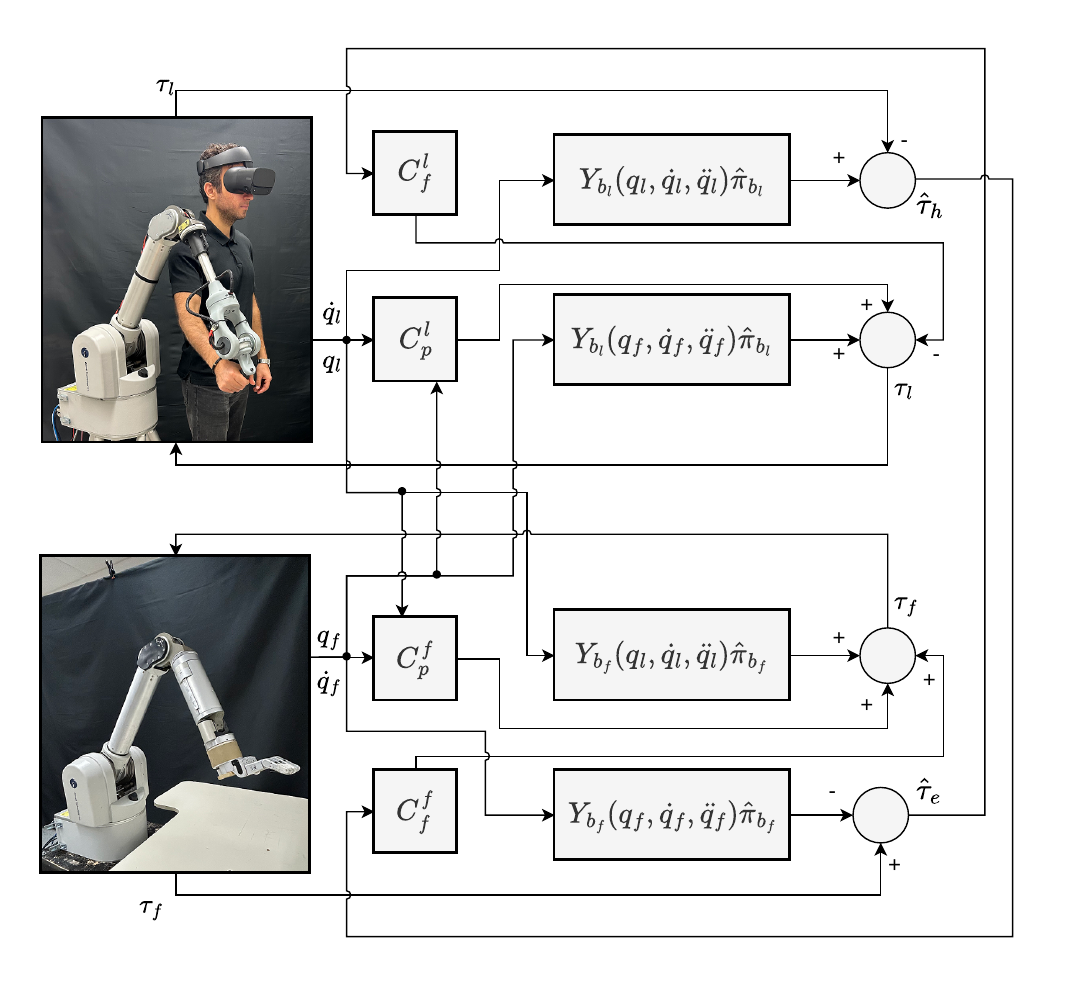}
        \caption{Schematic of the four-channel control architecture on the WAM bilateral teleoperation system.}
        \label{fig: wamscheme}
        \vspace{-15pt}
\end{figure}

Fig. \ref{fig: wamscheme} provides a schematic of the proposed sensorless four-channel controller on the WAM teleoperation system. Details regarding gain tuning for passivity are provided in Section~\ref{experiment:passivity}.

\section{EXPERIMENTS}

\subsection{Experiment Setup}

The teleoperation system consists of a 7-DOF WAM Barrett arm as the follower and a 4-DOF WAM Barrett arm equipped with a 3-DOF custom haptic wrist as the leader (Fig.~\ref{fig:wam teleop setup}). The wrist mirrors the kinematics of the WAM wrist, providing intuitive orientation feedback. The cable-driven WAM design provides high back-drivability, low friction, and relatively low moving mass, characteristics that have previously been shown to enable good transparency even with simple two-channel P-P control using only gravity compensation \cite{glover2009effective}.

The control software was developed in C++ using the open-source \texttt{libbarrett} library for low-level access to the WAM arms. External control mode was used to implement custom teleoperation loops and integrate the proposed architecture. Communication between leader and follower was established via UDP on the same computer, minimizing delay. The controller runs at 500~Hz, ensuring stable bilateral operation.

\subsection{Inverse Dynamics Estimation}\label{experiment:inverse}

In this section, the base dynamic parameters of the leader and follower WAM robots are identified. Wrist dynamics are neglected due to their relatively small contribution; thus, the wrists are modeled as rigid links and only the first four DOFs are considered. The leader and follower parameters are estimated independently using the open-source implementation in \cite{sousa2013physically}.

Each robot is excited using the optimal trajectory described in Section~\ref{inversedynamics}. Joint positions, velocities, and torques are recorded, while accelerations are obtained via central differencing of velocities and filtered to reduce noise. The base parameters are then estimated following Section~\ref{inversedynamics}. For a 4-DOF WAM, 26 base parameters are identifiable.

To validate the identified parameters, five random reference trajectories are executed on each robot. Model accuracy is evaluated using the Normalized Root
Mean Square Error (NRMSE) between the measured torques \(\tau_{ref}\) and the predicted torques \(\tau_{est} = Y_b(q_{ref}, \dot{q}_{ref}, \ddot{q}_{ref}) \hat{\pi}_b\). Validation is performed separately for the leader and follower, and results are reported in Table~\ref{table:1}. The low NRMSE values confirm that the identified models accurately capture the robot dynamics.

\begin{table}[t]
\caption{NRMSE (mean (std), \%) between the measured torques and the estimated torques from the base parameters.}
\label{table:1}
\begin{center}
\begin{tabular}{c|c|c|}
\cline{2-3}
                              & Leader WAM arm & Follower WAM arm \\ \hline
\multicolumn{1}{|c|}{Joint 1} & 5.25 (0.34)    & 5.34 (1.07)      \\ \hline
\multicolumn{1}{|c|}{Joint 2} & 2.31 (0.20)    & 2.79 (0.55)      \\ \hline
\multicolumn{1}{|c|}{Joint 3} & 5.69 (2.50)    & 5.46 (0.72)      \\ \hline
\multicolumn{1}{|c|}{Joint 4} & 3.69 (0.39)    & 3.01 (0.34)       \\ \hline
\end{tabular}
\end{center}
\end{table}

\subsection{Experimental Modeling and Gain Tuning for Passivity}\label{experiment:passivity}

Beyond transparency, stability and passivity are critical for safe teleoperation. To avoid energy injection, we experimentally evaluated the impact of modeling assumptions and controller settings, and tuned the parameters in (\ref{eq:26}) accordingly. Communication delays were not considered, as both robots were executed on the same computer and the focus was on modeling-related stability. The stability and passivity properties reported here are supported by experimental evaluation under the tested operating conditions.

Three observations guided the tuning process. First, including stiction and Coulomb friction degraded passivity by injecting energy; therefore, only viscous friction was retained (Section~\ref{inversedynamics}). Second, raw joint accelerations introduced instability due to noise, whereas scaling them to 25\% ensured stable and passive behavior. Third, full force feedback ($C_f=1$) caused instability; a reduced gain ($C_f=0.5$) preserved passivity with acceptable transparency. These modeling and gain selections reflect a practical stability--transparency trade-off suitable for typical human-scale manipulation tasks, which involve moderate accelerations rather than highly dynamic or fine micro-manipulation.

Table~\ref{tab:gains} lists the position and force gains used in the experiments. Identical gains were applied to both leader and follower. Only the first four DOFs were controlled; the wrist joints were locked and modeled as a rigid link.

\begin{table}[b]
\caption{Position control and force feedback gains for the first four joints of the WAM robots.}
\centering
\begin{tabular}{|c|c|c|c|c|}
\hline
Joint $i$ & 1 & 2 & 3 & 4 \\ \hline
$k_{p,i}$ & 750 & 1000 & 400 & 200 \\ \hline
$k_{i,i}$ & 2.5 & 1 & 2.5 & 0.5 \\ \hline
$k_{d,i}$ & 8.3 & 8 & 3.3 & 0.8 \\ \hline
$C_{f,i}$ & 0.5 & 0.5 & 0.5 & 0.5 \\ \hline
\end{tabular}
\label{tab:gains}
\end{table}

\begin{figure}[t]
  \centering
  \subfloat[]{\includegraphics[width=0.38\columnwidth]{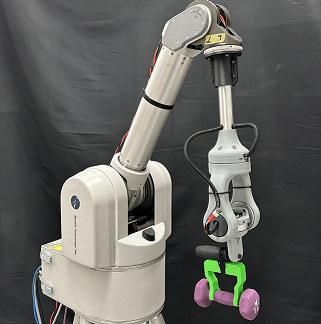}\label{fig:a}}
  \hspace{0.02\columnwidth}
  \subfloat[]{\includegraphics[width=0.38\columnwidth]{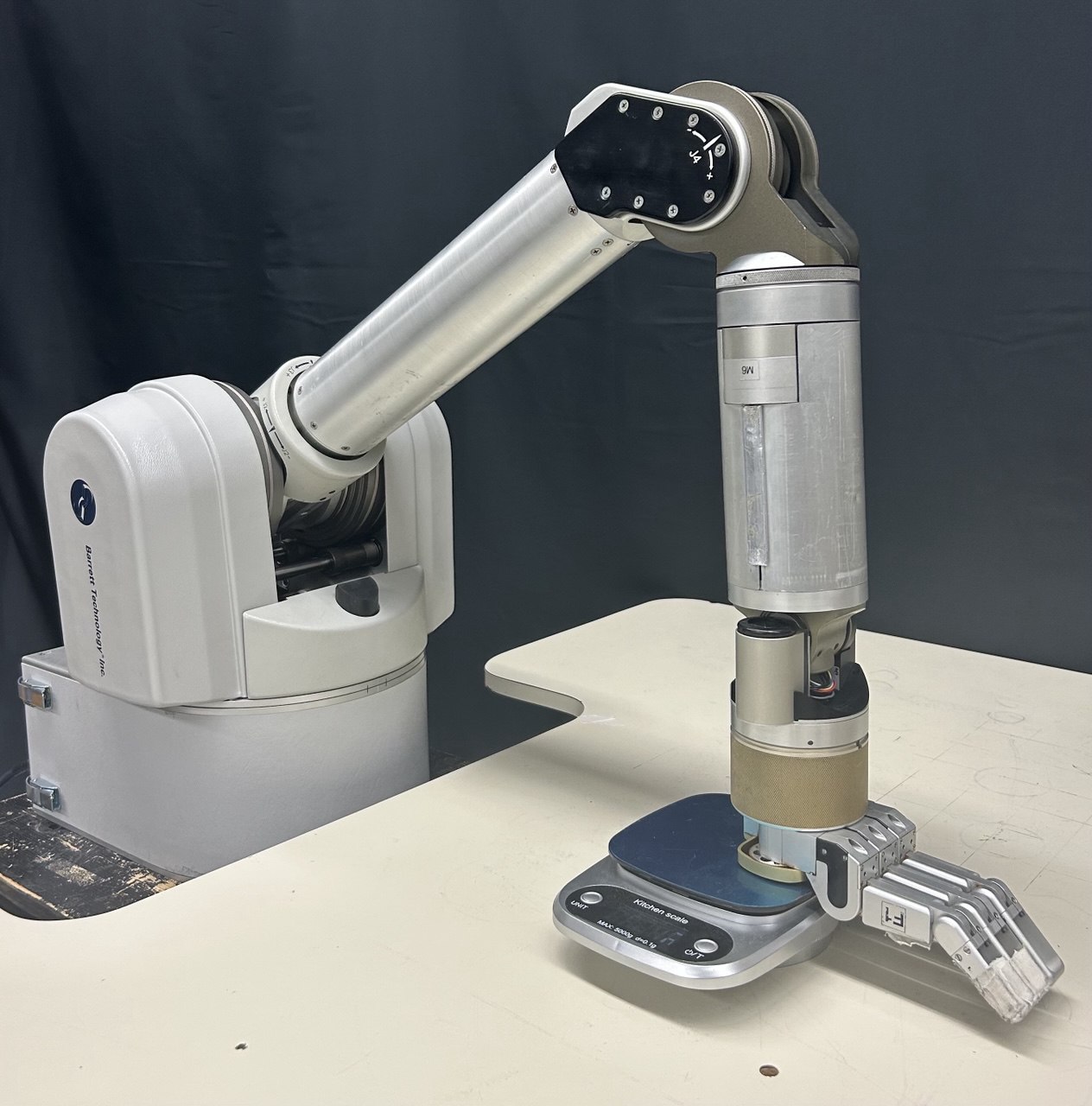}\label{fig:b}}
  \caption{Evaluation experimental setup: (a) weights mounted on the leader arm’s end-effector and (b) follower arm interacting with the kitchen scale.}
  \label{fig:experiment}
\end{figure}

\subsection{Evaluation Procedure of Control Architectures}\label{evaluationprocedure}

We follow the experimental transparency evaluation framework of \cite{aliaga2004experimental}, described in Section~\ref{transparencyanalysis}. The evaluation is conducted in joint space, consistent with the controller implementation, and all experiments are performed without a human operator.

\textit{Free motion:} Known feedforward joint torques are applied to the leader, and the resulting joint positions are recorded. Tracking performance is quantified using the NRMSE between leader and follower joint positions, which should ideally approach zero. Leader impedance is computed as $\hat{\tau}_{h,\mathrm{RMS}} / \delta q_{l,\mathrm{RMS}}$, where RMS denotes the root mean square and $\delta q_l$ the change in leader position; ideally, it approaches zero. Two torque magnitudes are applied three times each, yielding six trials.

\textit{Hard contact:} Calibrated weights are mounted on the leader end-effector (Fig.~\ref{fig:experiment}(a)), while the follower end-effector is pressed against a kitchen scale with 0.1~g resolution (Fig.~\ref{fig:experiment}(b)). The robots are configured such that the last link is perpendicular to the ground, ensuring normal contact. Force tracking is evaluated using the NRMSE between the estimated external joint torques of leader and follower, ideally approaching zero. The maximum transmittable impedance is computed as $\hat{\tau}_{h,\mathrm{RMS}} / \mathrm{RMSE}(q_l, q_f)$, which ideally tends to infinity. Two weight values are tested three times each, resulting in six trials.

\textit{External torque estimation:} The ground-truth leader force is given by the mounted weights, and the follower force by the scale measurement. Using (\ref{eq:222}), Cartesian forces are mapped to joint torques via the Jacobian transpose. The estimation error is quantified by the NRMSE between estimated and ground-truth joint torques.

Due to the WAM kinematic structure, Joints~2 and~4 contribute most to the workspace and are more sensitive to dynamic variations, particularly in contact tasks. Therefore, results are reported only for these joints for clarity.

\begin{figure}[t]
  \centering
  \includegraphics[width=\columnwidth]{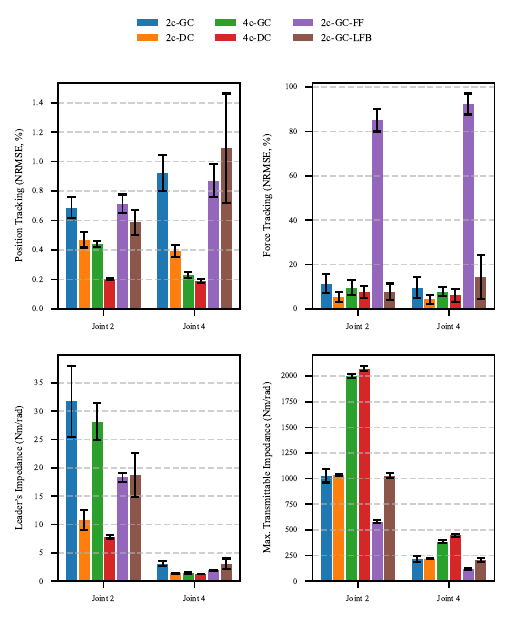}
  \caption{Transparency performance metrics across different teleoperation systems. Metrics: free motion position tracking, leader's impedance, hard contact force tracking, and maximum transmittable impedance.}
  \label{fig:perf-bars}
\end{figure}

\subsection{Comparison of Different Teleoperation Systems}\label{experiment:comparison}

The objective evaluation in Section~\ref{evaluationprocedure} is used to compare the proposed four-channel system (4c-DC) with three baseline controllers: two-channel P–P with gravity compensation (2c-GC) \cite{glover2009effective}, two-channel P–P with dynamics compensation (2c-DC), and four-channel with gravity compensation (4c-GC). Gravity compensation is included in all baselines, as a pure PID controller would be impractical for heavy arms. The 2c-DC controller isolates the effect of dynamics compensation, while 4c-GC isolates the effect of force feedback.

Fig.~\ref{fig:perf-bars} summarizes the results. Moving from 2c-GC to 2c-DC significantly improves free-motion tracking and leader impedance ($p<0.05$), confirming the benefit of dynamics compensation, while hard-contact force tracking and maximum transmittable impedance remain unchanged. Comparing 4c-GC with 2c-GC significantly increases the maximum transmittable impedance ($p<0.05$), demonstrating the role of force feedback in hard contact. Notably, 4c-GC also improves free-motion tracking and leader impedance. Overall, 4c-DC outperforms all baseline controllers in free-motion position tracking, leader impedance, and maximum transmittable impedance ($p<0.05$). Force tracking does not differ significantly across controllers, likely because the leader and follower are nearly identical and use identical gains.

In addition to the baseline architectures, we further compare 4c-DC with transparency-enhancement methods used in human-scale teleoperation: 2c-GC-FF and 2c-GC-LFB, which apply force feedforward \cite{fahmi2018inertial} and local force feedback \cite{hastrudi1999use} on the leader, respectively, on top of 2c-GC. A gain of 0.75 is used for both, as higher gains caused instability. As shown in Fig.~\ref{fig:perf-bars}, relative to 2c-GC, both methods improve leader impedance, enhancing free-motion transparency. In hard contact, however, 2c-GC-FF degrades maximum transmittable impedance and force tracking, while 2c-GC-LFB provides no significant improvement. In both free motion and hard contact, 4c-DC achieves superior transparency.

\begin{table}[t]
\caption{NRMSE between estimated and ground-truth external joint torques (mean (std), \%).}
\label{table:ext_error}
\begin{center}
\setlength{\tabcolsep}{4pt}
\begin{tabular}{c|c|c|c|c|}
\cline{2-5}
 & \multicolumn{2}{c|}{Leader WAM arm} 
 & \multicolumn{2}{c|}{Follower WAM arm} \\ \hline
\multicolumn{1}{|c|}{Controller} 
 & Joint 2 & Joint 4 & Joint 2 & Joint 4 \\ \hline
\multicolumn{1}{|c|}{2c-GC} 
 & 4.19 (2.83) & 3.38 (2.13) 
 & 11.14 (7.97) & 10.99 (4.45) \\ \hline
\multicolumn{1}{|c|}{2c-DC} 
 & 5.44 (1.27) & 3.77 (2.83) 
 & 5.95 (12.69) & 9.54 (3.79) \\ \hline
\multicolumn{1}{|c|}{4c-GC} 
 & 3.41 (1.97) & 6.58 (2.20) 
 & 8.31 (4.73) & 5.83 (4.91) \\ \hline
\multicolumn{1}{|c|}{4c-DC} 
 & 3.23 (2.18) & 8.18 (4.63) 
 & 11.72 (8.62) & 9.93 (7.21) \\ \hline
\multicolumn{1}{|c|}{2c-GC-FF} 
 & 6.13 (8.07) & 4.83 (3.46) 
 & 1.22 (1.10) & 2.62 (0.63) \\ \hline
\multicolumn{1}{|c|}{2c-GC-LFB} 
 & 3.19 (3.15) & 2.84 (0.11) 
 & 12.71 (7.74) & 5.11 (1.81) \\ \hline
\end{tabular}
\end{center}
\end{table}

Table~\ref{table:ext_error} reports the external torque estimation error for leader and follower. The mean estimation errors remain small across all controllers.


\begin{figure}[t]
  \centering
  \subfloat[]{\includegraphics[width=0.35\columnwidth]{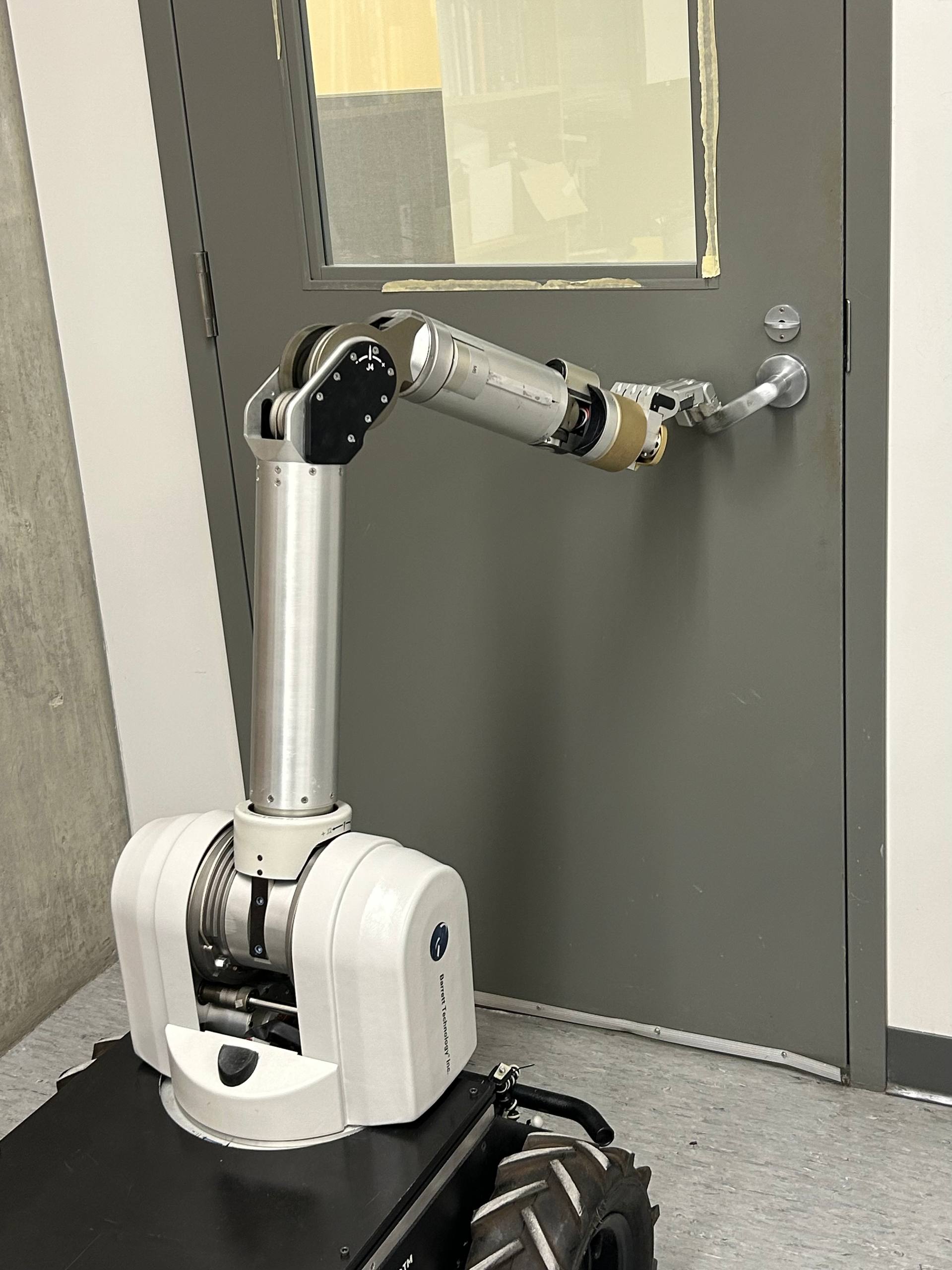}\label{fig:a}}
  \hspace{0.02\columnwidth}
  \subfloat[]{\includegraphics[width=0.35\columnwidth]{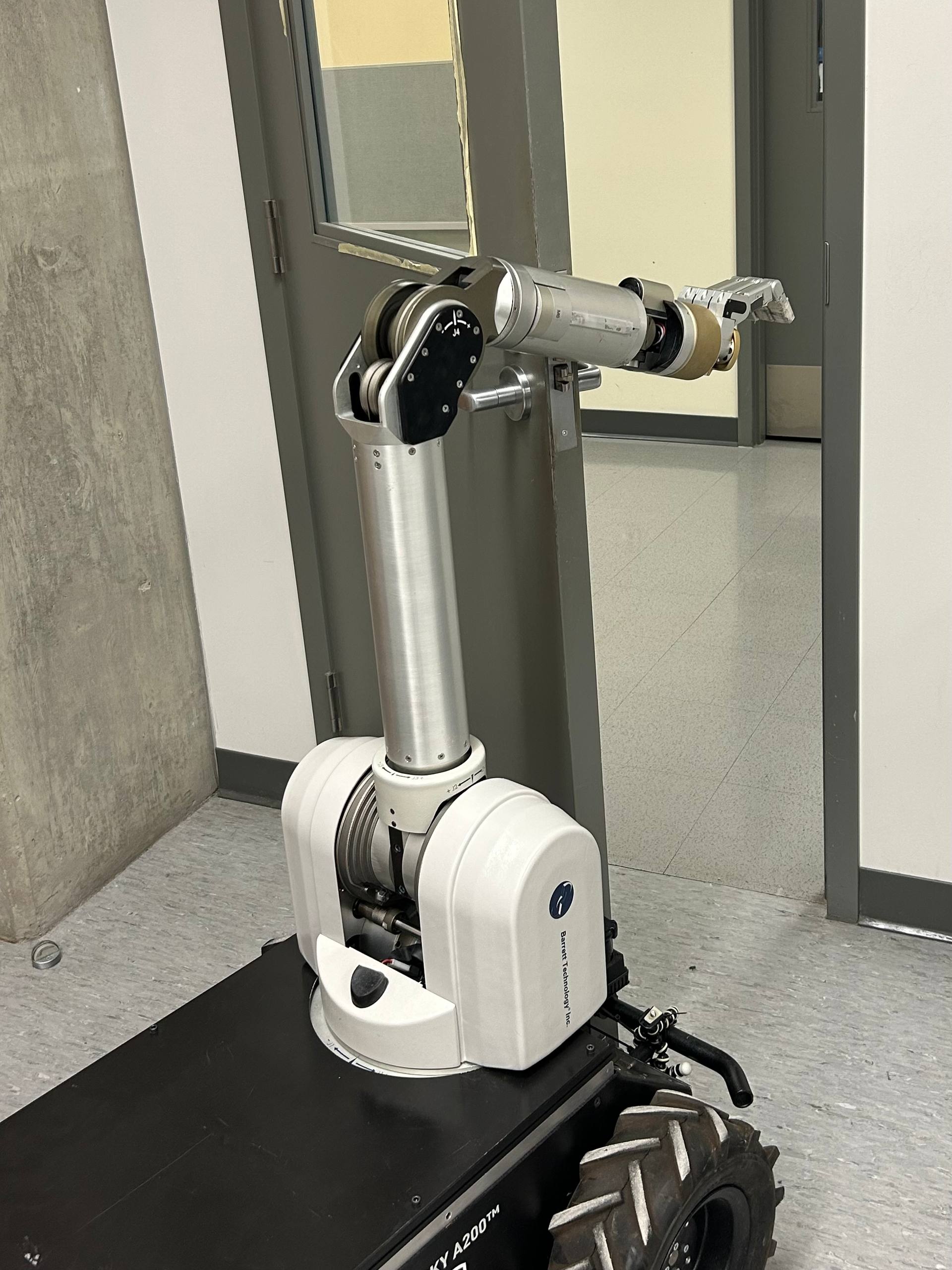}\label{fig:b}}
  \caption{Door-opening task: (a) follower arm manipulating the door handle; (b) whole-body contact along the follower arm while pushing the door open.}
  \label{fig:door_task}
\end{figure}

\begin{figure*}[t]
  \centering
  \includegraphics[width=\textwidth]{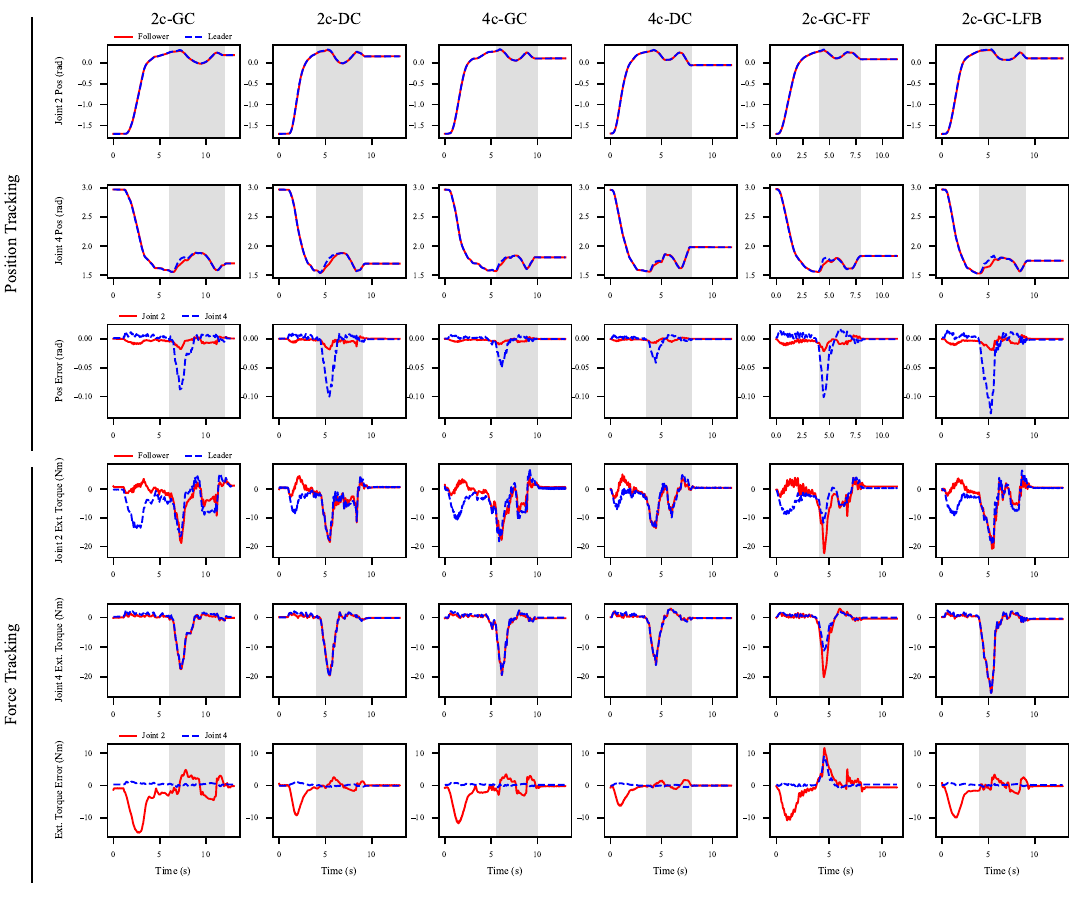}

  \caption{Comparison of teleoperation systems for the door-opening task using joint positions for position tracking and external joint torques for force tracking. Gray regions indicate the contact period.}

  \label{fig:compression_task}
\end{figure*}






\subsection{Case Study: Door-Opening Task}

We evaluate the proposed four-channel architecture on a door-opening task. A human operator uses the leader to drive the follower arm to manipulate the door handle and push the door open (Fig.~\ref{fig:door_task}). During the task, interactions occur both at the end-effector and along the follower arm, in addition to the kinematic constraints imposed by the door hinge. The proposed system is compared with the controllers described in Section~\ref{experiment:comparison}.

Fig.~\ref{fig:compression_task} shows position and force tracking for all methods. The gray region indicates the contact period during which the follower applies force to open the door. The proposed 4c-DC system outperforms all other methods in both free motion and hard contact. Improved position tracking during contact reflects higher maximum transmittable impedance, enabling better perception of door constraints, while improved force tracking in free motion indicates lower leader impedance and easier operator motion. Overall, 4c-DC achieves the highest transparency across both regimes.

\FloatBarrier

\addtolength{\textheight}{-4cm} 

\section{CONCLUSIONS}

This work presented a sensorless four-channel teleoperation architecture based on inverse dynamics modeling to enhance transparency in human-scale teleoperation. Experiments on a WAM bilateral teleoperation system showed that the proposed method improves position and force tracking, reduces operator effort in free motion, and increases the maximum transmittable impedance in hard contact, enabling perception of contact interactions occurring both at the end-effector and along the manipulator body.

For future work, improving the modeling of unmodeled dynamics, particularly stiction and Coulomb friction, could further enhance transparency, especially during slow motions and fine manipulation. Incorporating online adaptation, for example through disturbance observers or adaptive inverse dynamics, could also compensate for modeling errors and parameter uncertainties in real time. Beyond model refinement, extending the framework to heterogeneous leader-follower systems through independent dynamic identification would broaden its applicability. Finally, evaluating the proposed architecture under communication delays will be important for assessing its robustness in practical teleoperation scenarios.









\bibliographystyle{IEEEtran}
\bibliography{IEEEabrv,references}

\end{document}